\documentclass[10pt,twocolumn,letterpaper]{article}

\usepackage{wacv}              

\usepackage{graphicx}
\usepackage{amsmath}
\usepackage{amssymb}
\usepackage{booktabs}
\usepackage{arydshln}
\usepackage{multirow}
\usepackage{enumitem}
\usepackage{algorithm}
\usepackage[noend]{algorithmic}
\usepackage[accsupp]{axessibility} 
\usepackage[pagebackref,breaklinks,colorlinks]{hyperref}

\usepackage[capitalize]{cleveref}
\crefname{section}{Sec.}{Secs.}
\Crefname{section}{Section}{Sections}
\Crefname{table}{Table}{Tables}
\crefname{table}{Tab.}{Tabs.}

\def\wacvPaperID{473} 
\def\confName{WACV}
\def\confYear{2025}

\begin{document}

\title{QuantAttack: Exploiting Quantization Techniques\\to Attack Vision Transformers}

\author{
Amit Baras$^*$, Alon Zolfi$^*$, Yuval Elovici, Asaf Shabtai\\
Ben-Gurion University of the Negev, Israel\\
\texttt{\small \{barasa,zolfi\}@post.bgu.ac.il, \{elovici,shabtaia\}@bgu.ac.il
} \\
}
\maketitle
\def\thefootnote{*}\footnotetext{Equal contribution}\def\thefootnote{\arabic{footnote}}

\begin{abstract}
\vspace{-0.25cm}
In recent years, there has been a significant trend in deep neural networks (DNNs), particularly transformer-based models, of developing ever-larger and more capable models.
While they demonstrate state-of-the-art performance, their growing scale requires increased computational resources (e.g., GPUs with greater memory capacity).
To address this problem, quantization techniques (i.e., low-bit-precision representation and matrix multiplication) have been proposed.
Most quantization techniques employ a static strategy in which the model parameters are quantized, either during training or inference, without considering the test-time sample.
In contrast, dynamic quantization techniques, which have become increasingly popular, adapt during inference based on the input provided, while maintaining full-precision performance.
However, their dynamic behavior and average-case performance assumption makes them vulnerable to a novel threat vector -- adversarial attacks that target the model's efficiency and availability.
In this paper, we present \mbox{QuantAttack}, a novel attack that targets the availability of quantized vision transformers, slowing down the inference, and increasing memory usage and energy consumption.
The source code is available online\footnote{\url{https://github.com/barasamit/QuantAttack}}.
\end{abstract}
\vspace{-0.4cm}
\vspace{-0.35cm}
\section{\label{sec:intro}Introduction}
\vspace{-0.08cm}
In recent years, deep neural networks (DNNs), particularly transformers~\cite{lin2022survey}, have made tremendous progress in various domains, such as NLP~\cite{kalyan2021ammus} and computer vision~\cite{khan2022transformers}.
Their success stems mainly from their continually expanding network size, which is based on the number of parameters they contain and the precision of the parameters (the number of bits with which each parameter is represented).
However, the high computational cost of their transformer blocks makes them unsuitable for edge devices.

One way of reducing the size of the parameters is to quantize them to fewer bits and use low-bit-precision matrix multiplication.
There are two types of quantization techniques: 
(a) quantization-aware training~\cite{vision-QAT,vision-QAT2,vision-QAT3} - training a model with quantized parameters; and 
(b) post-training quantization~\cite{PTQ1-vision, PTQ2-vision,PTQ3-vision} - quantizing parameters of a pretrained model, which can be done in a static or dynamic (sample-dependent) manner.
Recently, LLM.int8()~\cite{LLM8}, a dynamic post-training quantization technique, was proposed and integrated into Hugging Face~\cite{wolf2019huggingface}, one of the largest open-source machine learning (ML) platforms.
This technique decomposes the features and weights into sub-matrices of large magnitude features (outliers) and other values.
The outlier feature matrices are multiplied in higher precision, while all other values are multiplied in lower precision, reducing inference time, memory usage, and energy consumption, without any degradation in performance.
While LLM.int8() was originally proposed for the NLP domain, the application of quantization techniques extends to vision models, aiming to balance efficiency and performance while preserving crucial visual information for tasks like image classification and object detection.
Despite the fact that quantization is effective in making DNNs more resource-efficient, it also opens up a new attack surface for adversaries aiming to compromise model availability. 

Given the potential impact of availability-oriented attacks, the ML research community has begun to direct its attention to adversarial attacks that target model availability.
Shumailov~\etal~\cite{shumailov2021sponge} were the first to present \textit{sponge examples}, a technique that exploits data sparsity, causing the number of GPU operations to increase; this leads to increased inference time and energy consumption.
Employing this attack vector, Cina~\etal~\cite{cina2022energy} poisoned models during the training phase to cause delays in the testing phase.
Dynamic neural networks~\cite{han2021dynamic}, which adapt their structures or parameters to the input during inference, have also been shown to be susceptible to adversarial attacks.
For example, adversarial attacks against models that employ early-exit strategies (a type of a dynamic network) have been studied extensively~\cite{multi-exit,pan2022gradauto}.
Research focusing on the post-processing phase of DNNs, particularly in object detection~\cite{shapira2023phantom} and LiDAR detection~\cite{liu2023slowlidar}, has shown that they are also susceptible to availability-oriented attacks.

In this paper, we introduce \emph{QuantAttack}, a novel adversarial attack that specifically targets the quantization process and exploits its test-time dynamism in vision transformers. 
We argue that this attack vector poses a significant threat to the availability of transformer models, as it operates at the core level and is broadly generalizable to any type of network.
Such attacks could have far-reaching implications, particularly in resource-constrained environments, where maintaining low latency and computational efficiency is paramount.
For instance, in cloud-based IoT systems like surveillance cameras, this attack can cause delays in anomaly detection, compromising security. 
Similarly, in real-time edge applications such as autonomous vehicles, the attack can lead to navigation errors or delayed decision-making, posing significant safety risks.
Above all, this paper aims to highlight the potential risk that dynamically quantized transformer-based models comprise.
To perform our attack, we propose overloading the matrix multiplication operation with high-bit precision (\ie, outlier values) to trigger worst-case performance.
To increase the stealthiness of our adversarial examples in cases where anomaly detection mechanisms are applied (\eg, monitoring shifts in the predicted distribution), we design our attack to preserve the model's original classification.
To assess the proposed attack's effectiveness, in our evaluation, we perform experiments addressing:
\textit{(i)} modality - investigating uni-modal and multi-modal; 
\textit{(ii)} model task - considering various computer vision applications; 
\textit{(iii)} attack variations - single-image, class-universal, and universal;
\textit{(iv)} transferability - examining whether the perturbations are transferable between different models.
Our experiments on the ViT model~\cite{VIT} show that the proposed attack can increase the model's use of GPU memory by 17.2\%, extend the GPU processing time by 9\%, and expand energy use by 7\%.
Finally, we demonstrate that quantized models, both static and dynamic, are susceptible to integrity-based attacks.

Our contributions can be summarized as follows:
\begin{itemize}[noitemsep,topsep=0pt,leftmargin=*]
    \item To the best of our knowledge, we are the first to identify dynamic quantization as a novel threat vector and propose an attack exploiting the availability of quantized models.
    \item We design a stealthy attack that preserves the model's original classification.
    \item We conduct a comprehensive evaluation on various configurations, examining different modalities and tasks, reusable perturbations, transferability, and ensembles.
    \item We shed light on the vulnerabilities and provide key insights into the security implications associated with the transformer architecture.
    \item We present various countermeasures that can be employed to mitigate the threat posed by our attack.
\end{itemize}

\vspace{-0.1cm}
\section{\label{sec:related}Related Work}
\vspace{-0.1cm}

\subsection{Quantization}
DNNs, particularly transformers, have achieved great success in various ML tasks~\cite{khan2022transformers,wolf2020transformers}. 
However, their computational complexity and large model size pose challenges for real-time applications and resource-limited devices. 
The size of a model is determined by the number of parameters and their precision.
The precision relates to the number of bits each weight value is stored with, typically 16 bits (also referred to as float16 or \textit{f16}) or 8 bits (also referred to as int8 or \textit{i8}).
To mitigate these challenges, techniques like quantization are employed.

Quantization is used to reduce the computational time and memory consumption of neural networks.
By quantizing the weights and activations into low-bit integers (\eg, a 16-bit float to an 8-bit integer), GPU memory usage can be reduced and inference can be accelerated, due to low-bit-precision matrix multiplication.
Two main quantization approaches have been proposed:
\begin{itemize}[noitemsep,leftmargin=*,topsep=0pt]
    \item \textbf{Quantization-Aware Training (QAT)}\cite{vision-QAT,vision-QAT2,vision-QAT3}: This method involves training the model with quantized weights and activations. 
    QAT maintains good performance, even when using low-precision formats.
    \item \textbf{Post-Training Quantization (PTQ)}: This approach takes a pretrained model and quantizes it directly, eliminating the need for extensive retraining, making it generally less computationally intensive compared to QAT.
    PTQ can be categorized into two main categories:
    \begin{itemize}[noitemsep,leftmargin=*,topsep=0pt]
        \item \emph{Static quantization} \cite{PTQ1-vision,PTQ2-vision,PTQ3-vision,PTQ4,bondarenko2021understanding,PTQ5,liu2021post,yuan2021ptq4vit}: The weights are quantized to lower precision only once using calibration sets, after the model is trained.
        \item \emph{Dynamic quantization} \cite{LLM8}: The weights and activations are quantized during runtime based on specific rules. 
    \end{itemize}
\end{itemize}

In this paper, we focus on PTQ techniques, highlighting that dynamic techniques possess an availability-based vulnerability due to their inherent dynamism, which uniquely impacts their security. 
Additionally, we demonstrate that both static and dynamic techniques are susceptible to integrity-based attacks.

\subsection{Availability Attacks}
Confidentiality, integrity, and availability, also known as the CIA triad, are a model that that typically serves as the basis for the development of security systems~\cite{samonas2014cia}.
In the context of DNNs, adversarial attacks targeting integrity~\cite{szegedy2013intriguing,goodfellow2014explaining,moosavi2016deepfool,moosavi2017universal,sitawarin2018darts,xu2020adversarial,zolfi2021translucent,zolfi2022adversarial} and confidentiality~\cite{behzadan2019adversarial,joud2021review} have received a great deal of research attention over the last few years.
However, adversarial attacks that target the availability of these models have only recently gained the attention of the ML research community.
Shumailov~\etal~\cite{shumailov2021sponge} were the pioneers in this area, introducing the sponge examples attack, a technique that primarily targets the efficiency of vision and NLP models. 
The authors propose to exploit: 
\textit{(i)} computation dimensions - expanding the internal representation size of inputs/outputs; and 
\textit{(ii)} data sparsity - forcing non-zero activations against the zero-skipping multiplications acceleration technique.
Both attacks lead to increased energy consumption and inference time.
Employing the data sparsity attack vector, Cina~\etal~\cite{sponge_poisoning} proposed sponge poisoning, a method aimed at degrading models' throughput by subjecting them to a sponge attack during the training phase.
Another noteworthy extension of the sponge examples (computation
dimension vulnerability) was presented by Boucher~\cite{boucher2022bad}, who introduced an adversarial attack on NLP models using invisible characters and homoglyphs, which, while undetectable to humans, can significantly affect the model's throughput.

Dynamic neural networks~\cite{han2021dynamic}, which optimize computational efficiency by adapting to input data during runtime, have also been shown to be vulnerable to adversarial attacks.
Haque~\etal~\cite{ILFO-attack} attacked DNNs that employ an layer-skipping mechanism by generating adversarial examples that go through all the layers.
Later, in a similar way, Hong~\etal~\cite{multi-exit} proposed an attack against early-exit mechanisms, generating malicious that bypass all early exits.
Pan~\etal~\cite{pan2022gradauto} proposed a unified formulation to construct adversarial samples to attack both the dynamic depth and width networks.

Another avenue of research, in which the post-processing phase of DNNs is targeted, has also been investigated.
Shapira~\etal~\cite{shapira2023phantom} showed that overloading object detection models by increasing the total number of candidates input into the non-maximum suppression (NMS) component can lead to increased execution times. 
Liu~\cite{liu2023slowlidar} extended this to LiDAR detection models.

In this paper, we propose a novel attack vector that has not been studied before -- an attack that targets the availability of dynamically quantized models.

\section{\label{sec:background}Background}

\subsection{Dynamic PTQ}
We focus on the one of the most popular PTQ techniques, LLM.int8()~\cite{LLM8}.
We consider a quantized model ${f: \mathcal{X}\to\mathbb{R}^M}$ that receives an input $x\in\mathcal{X}$ and outputs $M$ real-valued numbers that represent the model's confidence for each class $m \in M$, and contains $L_q$ quantized layers.
During inference, for every quantized layer, given the layer input ${\textbf{X}_{f16}\in\mathbb{R}^{s\times h}}$ and weights ${\textbf{W}_{f16}\in\mathbb{R}^{h\times o}}$ with sequence dimension $s$, feature dimension $h$, and output dimension $o$, the steps for efficient matrix multiplication are:
\begin{enumerate}[noitemsep,leftmargin=*,topsep=0pt]
    \item \textbf{Outlier Extraction}: From the input $\textbf{X}_{f16}$, extract all column indices that contain at least one outlier (\ie, absolute values that are larger than a certain threshold $\tau$) into the set $O = \{ i \; \vert \;\exists \vert\textbf{X}_{f16}^{i}\vert > \tau , 0 \ge i \ge h\}$.
    \item \textbf{Mixed-Precision Multiplication}: The matrix multiplication process is divided into two segments. 
    Outliers are multiplied using  the standard matrix multiplication in float16, while non-outliers are first quantized to their 8-bit representation and then multiplied in int8. 
    This involves row-wise quantization for the input and column-wise quantization for the weight matrix.
    \item \textbf{Dequantization and Aggregation}: The non-outlier results are dequantized back to float16 and combined with the outlier results to form the final output.
\end{enumerate}

More formally, the matrix multiplication can be described as: 
\begin{equation}
\textbf{C}_{f16} \approx \sum_{h \in O} \textbf{X}_{f16}^{h}  \textbf{W}_{f16}^{h} +  \textbf{S}_{f16} \cdot \sum_{h \notin O}  \textbf{X}_{i8}^{h}  \textbf{W}_{i8}^{h}
\end{equation}
where $C_{f16}$ represents the output tensor in float16, $(X_{f16}^{h}, W_{f16}^{h})$ represent the float16 input and weight for outliers, $S_{f16}$ is the denormalization term for int8 inputs and weights, and $(X_{i8}^{h}, W_{i8}^{h})$ represent the int8 input and weight for non-outliers.
Additional details on the quantization process can be found in the supplementary material.

\subsection{Vision Transformers}

Vision transformers are usually comprised of $L$ transformer blocks, each of which containing a multi-head self-attention (MSA) and a feedforward network (FFN).
Given an input sequence representation $\textbf{Z}_l$, The $l$-th block output $\textbf{Z}_l^{'}$ is calculated as follows:
\vspace{-0.1cm}
\begin{equation}
    \label{eq:block}
    \textbf{Z}_l^{'} = \hat{\textbf{Z}} + \text{FFN}(f_\text{LN1}(\hat{\textbf{Z}})), \hspace{5pt}\text{where}\hspace{5pt} \hat{\textbf{Z}} = \textbf{Z}_l + \text{MSA}(f_\text{LN2}(\textbf{Z}_l))
\end{equation}
where $f_\text{LN}$ denotes layer normalization such that:
\begin{equation}
    \label{eq:norm}
    {f_\text{LN}(x)=\frac{x-\mathrm{E}[x]}{\sqrt{\mathrm{Var}[x]+\epsilon}} \cdot \gamma + \beta}
\end{equation}
with learnable parameters $\gamma$ and $\beta$.
Specifically, two different normalization layers are used to in the transformer block differing in the $\gamma$ and $\beta$ values.

A single-head attention is formulated as follows:
\begin{equation}
    \text{Attn}(Q,K,V) = \text{softmax}(\frac{QK^T}{\sqrt{d_k}})V
\end{equation}
where Q, K, and V are the query, key and value matrices, respectively, and $d_k$ is scaling factor.
Consequently, the MSA is computed as:
\begin{equation}
    \begin{gathered}
        \text{head}_{n,l} = \text{Attn}(\textbf{Z}_l \textbf{W}^Q_{n,l},\textbf{Z}_l \textbf{W}^K_{n,l},\textbf{Z}_l \textbf{W}^V_{n,l})\\
        \text{MSA}(\textbf{Z}_l) = \text{Concat}(\text{head}_{1,l},\ldots,\text{head}_{H,l})\textbf{W}_l^O,
    \end{gathered}
\end{equation}
where $\textbf{W}^Q_{n,l}, \textbf{W}^K_{n,l}, \textbf{W}^V_{n,l}, \textbf{W}^O_{n,l}$ are the parameter matrices of the $n$-th attention head in the $l$-th transformer block.
Finally, as shown in Equation~\eqref{eq:block}, the output of the MSA is then processed by the FFN, a two-layer MLP.

In the context of LLM.int8(), which quantizes linear layers, we also discuss their location in the model.
In the standard transformer block (presented above), the MSA and the FFN consist of four and two linear layers, respectively, for a total of six linear layers.
In the MSA, three linear layers (with parameters $\textbf{W}^Q_{n,l}, \textbf{W}^K_{n,l}, \textbf{W}^V_{n,l}$) are used to derive the parameters matrices for the query, key and value; the last linear layer (with parameter $\textbf{W}^O_{n,l}$) is used to combine the different attention heads.
In the FFN, a two-layer MLP is used to produce the final block's output.
Notably, the inputs fed into the first three layers in the MSA ($\hat{\textbf{Z}}$) and the first layer in the FFN ($\textbf{Z}_l$) are directly affected by the normalization layers $f_\text{LN1}$ and $f_\text{LN2}$, an important aspect we will further discuss in Section~\ref{sec:disc}.
\vspace{-0.1cm}
\section{\label{sec:method}Method}
\vspace{-0.05cm}
\subsection{\label{subsec:method:threat_model}Threat Model}

\noindent\textbf{Adversary's Goals.}
We consider an adversary whose primary goal is to generate an adversarial perturbation $\delta$ that triggers the worst-case performance of dynamic quantization techniques, \ie, increases the number of high-bit operations.
Along with our primary goal, to increase the stealthiness of the attack the adversary aims to maintain the original classification.

\noindent\textbf{Adversary's Knowledge.}
To assess the security vulnerability of dynamic quantization to adversarial attacks, we consider three scenarios:
\textit{(i)} a white-box scenario: the attacker has full knowledge about the victim model;
\textit{(ii)} a grey-box scenario: the attacker has partial knowledge about the set of potential models; and
\textit{(iii)} a black-box scenario: the attacker crafts a perturbation on a surrogate model and applies it to a different victim model.

\noindent\textbf{Attack Variants.}
Given a dataset $\mathcal{D}$ that contains multiple pairs $(x_i,y_i)$ where $x_i$ is a sample and $y_i$ is the label, we consider three variants:
\textit{(i)} single - a different perturbation $\delta$ is crafted for each $x \in \mathcal{D}$;
\textit{(ii)} class-universal - a single perturbation $\delta$ is crafted for a target class $m \in M$; and
\textit{(iii)} universal - a single perturbation $\delta$ is crafted for all $x \in \mathcal{D}$.

\subsection{\label{sec:Quant_Attack}The Quant Attack}

To achieve the goals presented above, we modify the PGD attack~\cite{madry2017towards} with a novel loss function~\cite{shapira2023phantom,multi-exit}.
The update of the perturbation $\delta$ in iteration $t$ is formulated as follows:
\begin{equation}
    \delta^{t+1} = \prod_{{\vert\vert\delta\vert\vert}_{p}<\epsilon}(\delta^{t}+\alpha\cdot\text{sign}(\nabla_\delta \sum\limits_{(x,y)\in\mathcal{D'}}\mathcal{L}(x,y)))
\end{equation}
where $\alpha$ is the step size, $\prod$ is the projection operator that enforces ${\vert\vert\delta\vert\vert}_p<\epsilon$ for some norm $p$, and $\mathcal{L}$ is the loss function.
The selection of $\mathcal{D'}$ depends on the attack variant:
\textit{(i)} for the single-image variant, $\mathcal{D'}=\{(x,y)\}$;
\textit{(ii)} for the class-universal variant with a target class $m\in M$, $\mathcal{D'}=\{(x,y)\in\mathcal{D}|y=m\}$; and 
\textit{(iii)} for the universal variant, $\mathcal{D'}=\mathcal{D}$.
Next, we describe the proposed custom loss function, which consists of two components:
\begin{equation}
    \label{eq:loss}
    \mathcal{L} = 
    \mathcal{L}_{\text{quant}} + 
    \lambda \cdot \mathcal{L}_{\text{cls}}
\end{equation}
where \( \lambda \) is empirically determined using the grid search approach.
The two components are described below.

\noindent\textbf{Quantization Loss.}
This component aims to achieve our main goal, increasing the number of multiplications in 16-bit precision.
The number of multiplications in a higher precision level depends on the existence of an outlier value in each column in the input hidden state matrix.
Therefore, practically, we aim to produce at least one ``synthetic" outlier value in each column in this matrix.

Formally, let $\mathbf{X}_{l_{q}}\in\mathbb{R}^{s\times h}$ denote the input of the ${l_q\text{-th}}$ quantized layer (we omit the bit-precision notation for simplicity).
For each input matrix $\mathbf{X}_{l_{q}}$, we extract the top-$K$ values of each column, denoted as $\mathbf{X}_{l_{q}}^{\text{top-}K}\in\mathbb{R}^{K\times h}$, with the aim of pushing these values towards a target value $x_\text{target}$, such that $x_\text{target} > \tau$.
The loss for a single layer is defined as follows:
\begin{equation}
    \mathcal{L}_\text{single-q}(\mathbf{X}_{l_{q}}^{\text{top-}K}) = 
    \frac{1}{K  h}
    \sum\limits_{\substack{x_h\in \mathbf{X}_{l_{q}}^{\text{top-}K} \\ \vert x_h \vert <  \tau}}
    \bigl(
    \vert x_h \vert - x_\text{target}
    \bigl)^2
\end{equation}
Note that we only use values below the threshold to ensure that existing outlier values are not penalized by the loss function.

Finally, the loss for all $L_q$ layers is defined as:
\begin{equation}
    \label{eq:quant_loss}
    \mathcal{L}_\text{quant} = \sum_{l_q}^{L_q} \mathcal{L}_\text{single-q}(\mathbf{X}_{l_{q}}^{\text{top-}K})
\end{equation}

\noindent\textbf{Classification Loss.}
To increase the stealthiness of our attack, we aim to preserve the original classification of the input image.
Therefore, we include the classification loss component, which is defined as follows: 
\begin{equation}
    \mathcal{L}_{\text{cls}} = 
    \frac{1}{M} \sum\limits_{m=1}^{M} (f_m(x+\delta) - f_m(x))^2
\end{equation}
where $f_m$ denotes the score for class $m$.
\section{\label{sec:eval}Evaluation}

\subsection{\label{sec:eval_setup}Evaluation Setup}

\noindent\textbf{Models.}
We evaluated our attack on two state-of-the-art vision transformers:
\begin{itemize}[noitemsep,leftmargin=*,topsep=0pt]
    \item \textbf{Vision Transformer (ViT)~\cite{VIT}:} We use the \textit{base} size version, pretrained on ImageNet-21K, at a resolution of 224x224.
    The model is then finetuned on ImageNet-1K .
    Images are presented to the model as a sequence of fixed-size patches (resolution 16x16).
    \item \textbf{Data-efficient image Transformer (DeiT)~\cite{DEIT}:} We use the \textit{base} size version, pretrained and finetuned on ImageNet-1K, at a resolution of 224x224.
    Images are presented to the model as a sequence of fixed-size patches (resolution 16x16).
\end{itemize}

In the supplementary material, we show that the accuracy results for the quantized models are on par with those of the non-quantized ones.

\begin{table*}[t]
    \centering
    \scalebox{0.6}{
    \begin{tabular}{l|ccccc|ccccc}
    \hline\hline
        & \multicolumn{5}{c|}{\textbf{ViT}} & \multicolumn{5}{c}{\textbf{DeiT}} \\
        \textbf{Perturbation} & Memory [Mbits] & Energy [mJ] & Throughput [ms] & Accuracy & Outliers & Memory [Mbits] & Energy [mJ] & Throughput [ms] & Accuracy & Outliers \\
        \hline
        \multirow{2}{*}{Clean} & 713.315 & 299800 & 20.145 & \multirow{2}{*}{100\%} & 237.325 & 835.611 & 525005 & 28.095 & \multirow{2}{*}{100\%} & 1520.570 \\
              & \(1.00 \times\) & \(1.00 \times\) & \(1.00 \times\) & & \(1.00 \times\) & \(1.00 \times\) & \(1.00 \times\) & \(1.00 \times\) & & \(1.00 \times\) \\[3pt]
        \multirow{2}{*}{Random} & 712.493 & 296993 & 19.983 & \multirow{2}{*}{48\%} & 235.148 & 849.006 & 524692 & 28.193 & \multirow{2}{*}{68\%} & 1556.900 \\
               & \(0.998 \times\) & \(0.991 \times\) & \(0.992 \times\) &  & \(0.991 \times\) & \(1.016 \times\) & \(0.999 \times\) & \(1.003 \times\) &  & \(1.024 \times\) \\[3pt]
        \multirow{2}{*}{Sponge Examples~\cite{shumailov2021sponge}} & 720.096 & 297079 & 20.459 & \multirow{2}{*}{88\%} & 288.363 & 888.930 & 525175 & 28.820 & \multirow{2}{*}{58\%} & 2400.000 \\
                       & \(1.009 \times\) & \(0.991 \times\) & \(1.016 \times\) & & \(1.215 \times\) & \(1.063 \times\) & \(1.00 \times\) & \(1.026 \times\) & & \(1.57 \times\) \\
        \multirow{2}{*}{Standard PGD~\cite{madry2017towards}} & 725.792 & 305242 & 20.054 & \multirow{2}{*}{0\%} & 344.884 & 887.43 & 525013 & 28.580 & \multirow{2}{*}{0\%} & 1867.000 \\
                   & \(1.017 \times\) & \(1.018 \times\) & \(0.995 \times\) &  & \(1.453 \times\) & \(1.06 \times\) & \(1.00 \times\) & \(1.01 \times\) &  & \(1.22 \times\) \\
        \cdashline{1-11}\noalign{\vskip 0.5ex}
        \multirow{2}{*}{Single} & \textbf{836.264} & \textbf{320591} & \textbf{21.924} & \multirow{2}{*}{\textbf{90\%}} & \textbf{3989.995} & \textbf{1078.701} & \textbf{554133} & \textbf{30.620} & \multirow{2}{*}{\textbf{89\%}} & \textbf{8094.00} \\
       & \(\textbf{1.172} \times\) & \(\textbf{1.069} \times\) & \(\textbf{1.088} \times\) & & \(\textbf{16.812} \times\) & \(\textbf{1.291} \times\) & \(\textbf{1.055} \times\) & \(\textbf{1.090} \times\) & & \(\textbf{5.324} \times\) \\
        \multirow{2}{*}{Universal\footnotemark[1]} & 730.923 & 311153 & 20.759 & \multirow{2}{*}{73\%} & 736.292 & 939.350 & 539060 & 29.030 & \multirow{2}{*}{81\%} & 3429.3 \\
          & \(1.025 \times\) & \(1.037 \times\) & \(1.030 \times\) &  & \(3.102 \times\) & 1.124 & 1.026 & 1.033 &  & 2.255 \\[3pt]
        \multirow{2}{*}{Class-Universal\footnotemark[1]} & 790.815 & 311862 & 20.850 & \multirow{2}{*}{83\%} & 868.687 & 941.880 & 54303 & 29.078
        & \multirow{2}{*}{81\%} & 4123.785 \\
          & \(1.108 \times\) & \(1.040 \times\) & \(1.034 \times\) &  & \(3.665\times\) & \(1.128 \times\) & \(1.044 \times\) & \(1.035 \times\) &  & \(2.712 \times\) \\[3pt]

        \hline\hline
       
    \end{tabular}}
    
     \caption{Evaluation of ViT and DeiT models on different baselines and attack variations. 
    The bottom number represents the percentage change from the clean image performance. 
    Bold indicates superior results.
    \protect\footnotemark[1]Perturbations are trained with $\epsilon=\frac{32}{255}$.}
    \label{tab:modified_tab}
    \vspace{-0.2cm}
\end{table*}

\noindent\textbf{Datasets.}
In our evaluation, we use the ImageNet dataset~\cite{deng2009imagenet}, and specifically, the images from its validation set, which were not used to train the models described above.
For the single-image attack variant, we trained and tested our attack on 500 random images from various class categories.  
For the class-universal variant, we selected 10 random classes, and for each class we trained the perturbation on 250 images (\ie, $\vert\mathcal{D}\vert=250$) and tested them on a distinct set of 500 images from the same class.
Similarly, for the universal variant, we followed the same training and testing procedure, however from different class categories.

\noindent\textbf{Metrics.}
To evaluate the effectiveness of our attack, we examine the number of outlier multiplications, different hardware metrics, and the effect of the attack on the model's original task performance:
\begin{itemize}[noitemsep,leftmargin=*,topsep=0pt]
    \item\textbf{GPU Memory Consumption}: how much GPU memory the process uses.
    \item\textbf{GPU Throughput}: how long the GPU takes to perform calculations.
    \item\textbf{Energy Consumption}: the total energy usage of the GPU, with measurements obtained using the NVIDIA Management Library (NVML).
    \item\textbf{Number of Outliers}: represents the number of matrix multiplications done in 16-bit precision (see Section~\ref{sec:background}).
    \item\textbf{Accuracy}: the performance of the model on its original task. 
    We consider the model's prediction on the original images as the ground-truth label.
\end{itemize}

\noindent\textbf{Implementation Details.}
In our attack, we focus on $\ell_{\inf}$ norm bounded perturbations, and set ${\epsilon=\frac{16}{255}}$, a value commonly used in prior studies~\cite{mahmood2021robustness,naseer2021improving,wei2022towards,zhang2023transferable}.
The results for other $\epsilon$ values can be found in the supplementary material.
We use a cosine annealing strategy with warm restarts for the attack's step size $\alpha$, where the maximum and minimum values are $\frac{0.5}{255}$ and $10e-5$, respectively.
For the dynamic quantization threshold, we set $\tau=6$, as suggested in the original paper~\cite{LLM8}.
We set the target value of our attack $x_\text{target}$ to 70, the number of extracted values from each column $K$ to 4, and the weighting factor $\lambda$ to 50 as they empirically yielded the best results.
The results of the ablation studies we performed on the $K$, $\lambda$ and $x_\text{target}$ values can be found in the supplementary material.
The experiments are conducted on a GeForce RTX 3090 GPU. 
olgu

\subsubsection{Effectiveness of Adversarial Perturbations}
We compare our attack with four baselines: (a) clean - the original image with no perturbation, random - a randomly sampled perturbation from the uniform distribution $\mathcal{U}(-\epsilon,\epsilon)$, Sponge Examples~\cite{shumailov2021sponge} - an attack aimed at increasing the amount of non-zero activations, and standard PGD~\cite{madry2017towards} - the original PGD attack with the model's loss function (integrity-based attack).
Table~\ref{tab:modified_tab} presents the performance of the different perturbations on the ViT and DeiT models.
The analysis reveals that single-image perturbations substantially increase the GPU memory usage and processing time for both models compared to the baselines.
Specifically, for the ViT model, the single-image perturbations cause 1681\% more outliers than a clean image.
In terms of hardware metrics, they result in a 17.2\% increase in GPU memory usage, a 7\% increase in energy consumption, and an 9\% increase in GPU processing time, compared to clean images.
The sponge examples and the standard PGD however, do not incur any substantial effect both in terms of outliers and hardware metrics.
Interestingly, in most cases, the random perturbations lead to a degradation in performance compared to the clean images.
We hypothesize that random perturbations simply eliminate ``natural" outlier features (\ie, those that exist in clean images) due to the random noise added to the images.

\vspace{-0.35cm}
\subsubsection{Universal and Class-Universal Perturbations}
\vspace{-0.12cm}
We also investigate the effect of reusable perturbations, in which class-universal and universal perturbations (Section~\ref{subsec:method:threat_model}) are trained on one set of images and tested on a distinct holdout set.
The results presented in Table~\ref{tab:modified_tab} enable comparison of the perturbations' impact based on various metrics.
On the DeiT model, when compared to clean images, a universal perturbation results in a 12.4\% increase in GPU memory usage, 2.6\% increase in energy consumption, and 3.3\% increase in GPU processing time. 
Class-specific perturbations cause an 12.8\% increase in GPU memory, 4.4\% increase in energy consumption, and 3.5\% increase in GPU time, performing slightly better than the universal perturbation.
Note that in this case, we used $\epsilon=\frac{32}{255}$ as it is a more complex setting compared to the single variant.

Although universal and class-specific perturbations are less resource-exhaustive than single-image perturbations, they offer a distinct advantage. 
A universal perturbation vector is capable of affecting multiple images or an entire class of images, thereby providing an efficient mechanism for broad-spectrum adversarial attacks. 
This efficiency is especially advantageous in scenarios where the attacker aims to disrupt the model across multiple data points with minimal computational effort.

When examining the effect of these perturbations, we observed an interesting phenomenon: relaxing the requirement for imperceptibility (\ie, the noise bound $\epsilon$ is set at a high value) causes the perturbation to completely distort the visual appearance of the input image, visually resembling sheer noise.
This, in turn, creates a resource-intensive scenario which could be interpreted as a denial-of-service (DoS) attack, increasing GPU memory usage by 70\%, energy consumption by 12\%, and GPU time by 25\%.

\begin{figure}[t]
    \centering
    \begin{subfigure}{.48\linewidth}
        \centering
        \includegraphics[width=\linewidth]{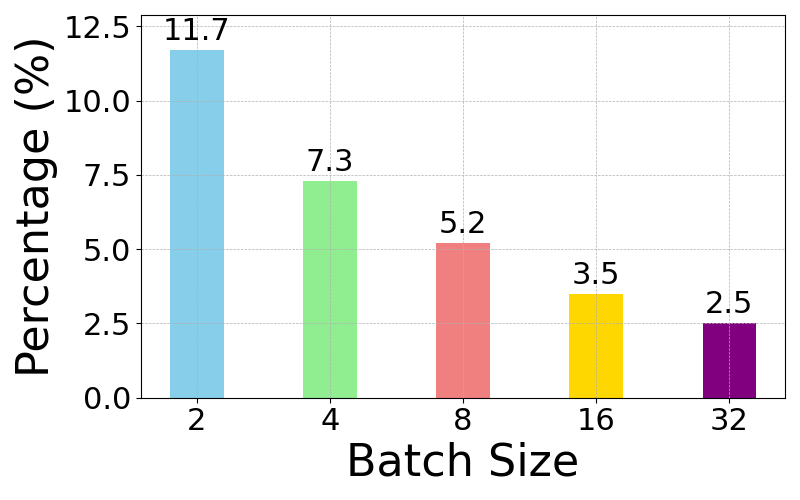}
        \caption{Memory Usage}
    \end{subfigure}
    \hspace{0.05cm}
    \begin{subfigure}{.48\linewidth}
        \centering
        \includegraphics[width=\linewidth]{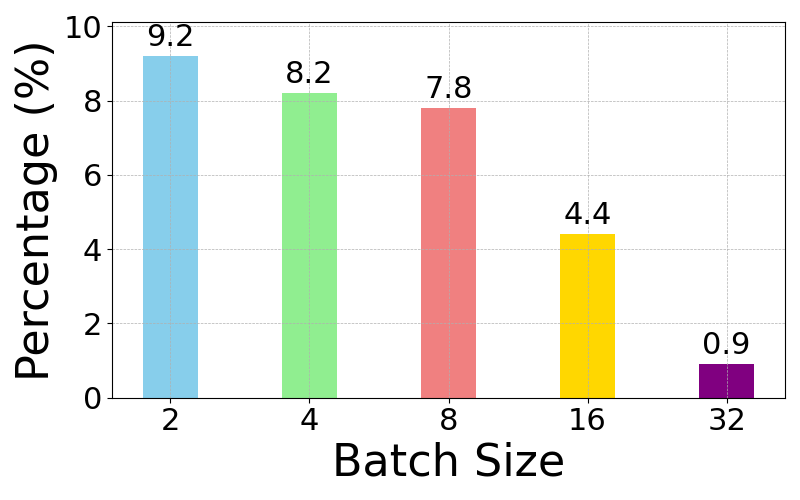}
        \caption{Processing Time}
        \label{subfig:batches:Time}
    \end{subfigure}
    \hspace{0.05cm}
    \begin{subfigure}{.48\linewidth}
        \centering
        \includegraphics[width=\linewidth]{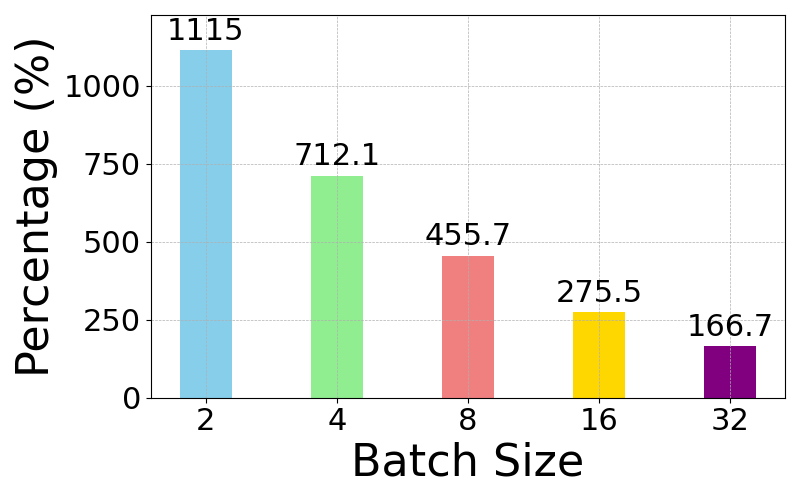}
        \caption{Number of Outliers}
        \label{subfig:batches:putliers}
    \end{subfigure}
    \hspace{0.05cm}
    \begin{subfigure}{.48\linewidth}
        \centering
        \includegraphics[width=\linewidth]{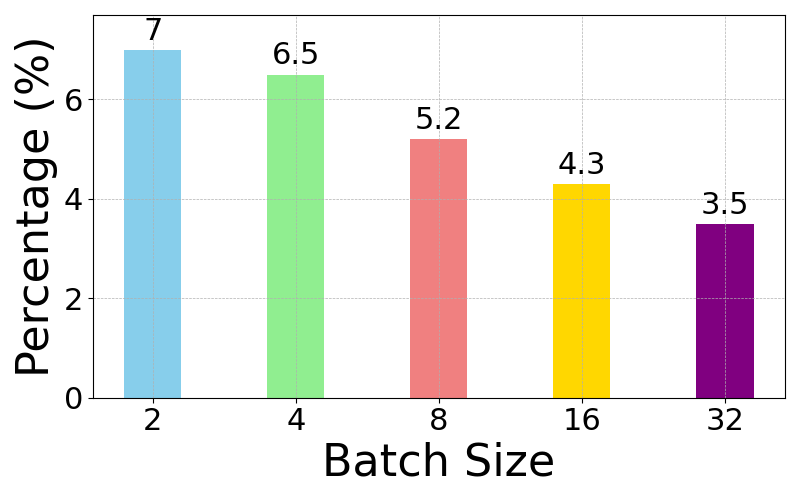}
        \caption{Energy Consump.}
        \label{subfig:batches:power}
    \end{subfigure}
    \caption{Evaluating the influence of a single perturbed image in a batch for different batch sizes.
    The values represent the percentage difference between a benign batch and its attacked counterpart.}
    \label{fig:batchs}
\end{figure}

\vspace{-0.1cm}
\subsubsection{Batch Attack}

In the LLM.int8()~\cite{LLM8} technique, when the quantized model processes a batch of $B$ images, every quantized layer transforms the given 3D input matrix ${\textbf{X}_{f16}\in\mathbb{R}^{B\times s\times h}}$ to a stacked 2D version ${\textbf{X}_{f16}\in\mathbb{R}^{(B\cdot s)\times h}}$, resulting in $B\times$ more rows.
The transformation is followed by quantized matrix multiplication (Section~\ref{sec:method}).
In this case, when an outlier value exists in a column, the entire column is processed in f16 precision, including values in rows that belong to a completely different image. 

In a realistic scenario where a quantized model is deployed and serves as ML-as-a-service, input images from different sources can be stacked together and given to the model as a batch of images.
In this case, an adversary could potentially implant a perturbed image that will affect the resource consumption of the whole batch.

Therefore, we evaluate the impact of such a scenario in which we feed the quantized model different sized batches that include a \emph{single} perturbed image.
Figure~\ref{fig:batchs} presents the results for the different batch sizes; the values represent the percentage difference between a benign batch and its attacked counterpart.
The results show that a single perturbed image implanted in a batch of images could potentially affect the performance of the entire batch, as an increase in resource consumption is seen for all of the batch sizes examined.
Notably, smaller size batches (\eg, two images) are more sensitive than larger size batches (\eg, 16 images).
For example, for a two-image batch, the memory usage increases by 11.7\% compared to the benign batch, while for a 16-image batch, the memory only increases by 3.5\%.
A plausible explanation for this can be found in the initial state of outliers in the batches - with a large batch size, natural outliers from different images are likely to spread across multiple columns.
Consequently, the attacked image could contain synthetic outliers in the same columns, which are already processed in f16, and this leads to a smaller percentage difference.

\vspace{-0.3cm}
\subsubsection{Transferability and Ensemble Approaches}

In adversarial attacks, transferability refers to the ability of an adversarial example, crafted on a surrogate model, to affect other models. 
In our experiments, we examine the effect of perturbations trained on ViT and tested on DeiT and vice versa.
As shown in Table~\ref{tab:Ensemble}, adversarial examples trained on one model show limited impact on the other.
Interestingly, we observed an unexpected anomaly in the GPU time when perturbations were trained on DeiT and tested on ViT, in which a negative value was recorded (\ie, the GPU time decreased).
We hypothesize that this occurred due to the marginal effect of just 10\% more outliers.
Such anomalies emphasize the nuanced relationship between outliers and resource metrics, hinting at the complex performance environment within which these models operate. 
Despite this minor deviation, the general trend remains consistent: a higher number of outliers usually requires more resources.

To improve the transferability between different models, we employ an ensemble training strategy, in which the adversarial example is trained on both models simultaneously, such that in each training iteration one of the models is randomly selected.
This approach aims to examine the collaborative benefit of utilizing the strengths of both models to create generalizable adversarial perturbations. 
Based on the results presented in Table~\ref{tab:Ensemble}, we can see that the ensemble strategy is able to affect both models, and results in an increase in GPU memory usage and GPU time, although with sightly less efficiency compared to perturbations trained and tested on the same model.
For example, when the ensemble-based perturbations are tested on the DeiT model, GPU memory usage and GPU time increased by 21.3\% and 7.5\%, respectively.

\subsubsection{Evaluating Across Diverse Model Architectures}
To emphasize the potential impact and generalizability of our attack, we also experiment with other models that have different characteristics: different computer vision tasks, and different modalities.
In addition, we also broaden our analysis from the computer vision domain to include the audio domain.
Particularly, we experiment with the following models:
(a) Open-World Localization (OWLv2)~\cite{minderer2023scaling} - a zero-shot text-conditioned object detection model;
(b) You Only Look at One Sequence (YOLOS)~\cite{fang2021you} - a transformer-based object detection model;
(c) Generative Image-to-text Transformer (GIT)~\cite{wang2021git} - a decoder-only transformer for vision-language tasks; and
(d) Whisper~\cite{Whisper} - a sequence-to-sequence model for automatic speech recognition and speech translation.
It should be noted that in our attack against those models, we only use the single-image attack variant with the quantization loss component (\ie, $\lambda=0$) for simplicity.
We also note that the same attack configuration used for the image classification models (Section~\ref{sec:eval_setup}) is used for all the evaluated models below, which might lead to suboptimal results.
Further experimental details on these models can be found in supplementary material.

From Table~\ref{tab:Other models} we can see that transformer-based models in general are vulnerable to our attack, not limited to the image classification domain.
For example, on the multi-modal OWLv2, the energy consumption increased by 4.9\% and on the YOLOS, the memory usage increased by 9.3\%.
Beyond the computer vision domain, our attack also successfully affects the Whisper model.

\begin{table}[t!]
\scalebox{0.78}{
\begin{tabular}{lcc|cc}
\toprule
& \multicolumn{2}{c|}{\textbf{Single Model}} & \multicolumn{2}{c}{\textbf{Ensemble}} \\ \cline{2-5}
\multicolumn{1}{r}{\textbf{Train Model(s)}} & \textbf{D} & \textbf{V}& \multicolumn{2}{c}{\textbf{D+V}} \\
\multicolumn{1}{r}{\textbf{Test Model}} & \textbf{V} & \textbf{D} & \textbf{V} & \textbf{D} \\
\midrule
\textbf{Memory} & 0.2\% & 1.7\% & 13.7\% & 21.3\% \\
\textbf{Throughput} & -0.02\% & 0\% & 7\% & 7.5\% \\
\textbf{Outliers} & 10\% & 11\% & 1344\% & 475\% \\
\bottomrule
\end{tabular}}
\caption{Transferability and ensemble analysis. 
Values indicate the percentage difference between perturbed and clean images.
D and V denote the DeiT and ViT models, respectively.}
\label{tab:Ensemble}
\end{table}

\begin{table}[t]
    \centering
    \scalebox{0.73}{
    \begin{tabular}{llcc}
    \toprule
    \textbf{Quantization} & \textbf{Attack} & \textbf{ViT} & \textbf{DeiT} \\ \midrule
    \multirow{2}{*}{PTQ4ViT~\cite{yuan2022ptq4vit}} & Ours & \textbf{10\%} & \textbf{10\%} \\ 
     & Random & 68\% & 68\% \\ \midrule
    \multirow{2}{*}{Repq-vit~\cite{li2023repq}} & Ours & \textbf{10\%} & \textbf{11\%} \\ 
     & Random & 62\% & 70\% \\ \midrule
    \multirow{2}{*}{LLM.int8()~\cite{LLM8}} & Ours & \textbf{26\%} & \textbf{20\%} \\
     & Random & 48\% & 68\% \\
    \bottomrule
    \end{tabular}}
    \caption{Accuracy comparison of ViT and DeiT models with various quantization techniques on the single-image attack variant.}
    \label{tab:comparison}
    \vspace{-0.3cm}
\end{table} 

\begin{table*}[t]
\centering
\resizebox{0.88\textwidth}{!}{%
\begin{tabular}{lcc|cc|cc|cc}
\hline
 & \multicolumn{2}{c}{\textbf{Memory [Mbits]}} & \multicolumn{2}{c}{\textbf{Energy [mJ]}} & \multicolumn{2}{c}{\textbf{Throughput [ms]}} & \multicolumn{2}{c}{\textbf{Outliers}} \\
\textbf{Model} & Clean & Single & Clean & Single & Clean & Single & Clean & Single \\
\hline
OWLv2~\cite{minderer2023scaling} & 24828.590 \((1.00 \times\)) & 25586.340 \((1.031 \times\)) & 1223276 \((1.00 \times\)) & 1256220 \((1.027 \times\)) & 217.886 \((1.00 \times\)) & 228.545 \((1.049 \times\)) & 8185.395 \((1.00 \times\)) & 18800.86 \((2.297 \times\)) \\[3pt]

YOLOS~\cite{fang2021you} & 13294.808 \((1.00 \times\)) & 13428.000 \((1.01 \times\)) & 519451 \((1.00 \times\)) & 567790 \((1.093 \times\)) & 163.89 \((1.00 \times\)) & 169.540 \((1.034 \times\)) & 386.988 \((1.00 \times\)) & 2656.144 \((6.864 \times\)) \\[3pt]

GIT~\cite{wang2021git} & 4139.994 \((1.00 \times\)) & 4233.725 \((1.023 \times\)) & 837583 (\(1.00 \times\)) & 856982 (\(1.023 \times\)) & 291.586 (\(1.00 \times\)) & 292.142 (\(1.002 \times\)) & 3407.774 \((1.00 \times\)) & 4990.700 \((1.465 \times\)) \\[3pt]

Whisper~\cite{Whisper} & 1347.759 \((1.00 \times\)) & 1377.000 \((1.022 \times\)) & 180104 \((1.00 \times\)) & 185466 \((1.029 \times\)) & 18.911 \((1.00 \times\)) & 19.177 \((1.014 \times\)) & 1651.250 \((1.00 \times\)) & 2793.000 \((1.692 \times\)) \\
\hline
\end{tabular}
}
\caption{Comprehensive evaluation of the performance of QuantAttack (single-image variant) on various computer vision and audio models. The number in parentheses represents the percentage change from the clean images' performance.}
\label{tab:Other models}
\vspace{-0.3cm}
\end{table*}

\vspace{-0.3cm}
\subsubsection{Integrity-Based Attacks}

Beyond the scope of availability-based attacks, we also explore the effect of generating outlier values on the models' prediction capabilities (\ie, use Equation~\ref{eq:loss} with $\lambda=0$).
We evaluate the effectiveness of our attack on both static and dynamic PTQ techniques. 
We argue that dynamic techniques heavily rely on outlier values for successful classification, while static techniques depend on low-scale calibration sets that do not account for extreme cases (\eg, artificially crafted outlier values).

We evaluate the models' performance under various quantization techniques.
Specifically, in addition to LLM.int8(), we also evaluate the static quantization methods PTQ4ViT~\cite{yuan2022ptq4vit} and Repq-vit~\cite{li2023repq}. 
As shown in Table~\ref{tab:comparison}, our attack successfully degrades model's accuracy, regardless of the quantization technique. 
Interestingly, the static techniques are more vulnerable, as they are not calibrated to handle extreme values. 
On the other hand, the dynamic technique is relatively more robust to outlier values due to its increased bit-precision mechanism. 
This highlights the trade-off between robustness to availability attacks and maintaining model performance.
We include transferability and ensemble results in the supplementary material.
\subsection{\label{sec:disc}Discussion and Insights}

In Section~\ref{sec:background} we discussed the presence of the normalization layers in the transformer block, and which linear layers' inputs are directly affected by them.
In the context of our attack, which aims to increase the values in these inputs, the normalization applied to the inputs just before the mixed-precision matrix multiplication has both negative and positive effects on our attack's performance. 
Given an input matrix $x$, the output of the normalization layer (Equation~\eqref{eq:norm}) is affected by the input mean $\mathrm{E}[x]$, variance $\mathrm{Var}[x]$, and the learnable parameters $\gamma$ and $\beta$.

\begin{figure}[t!]
    \centering
    \includegraphics[width=1\linewidth]{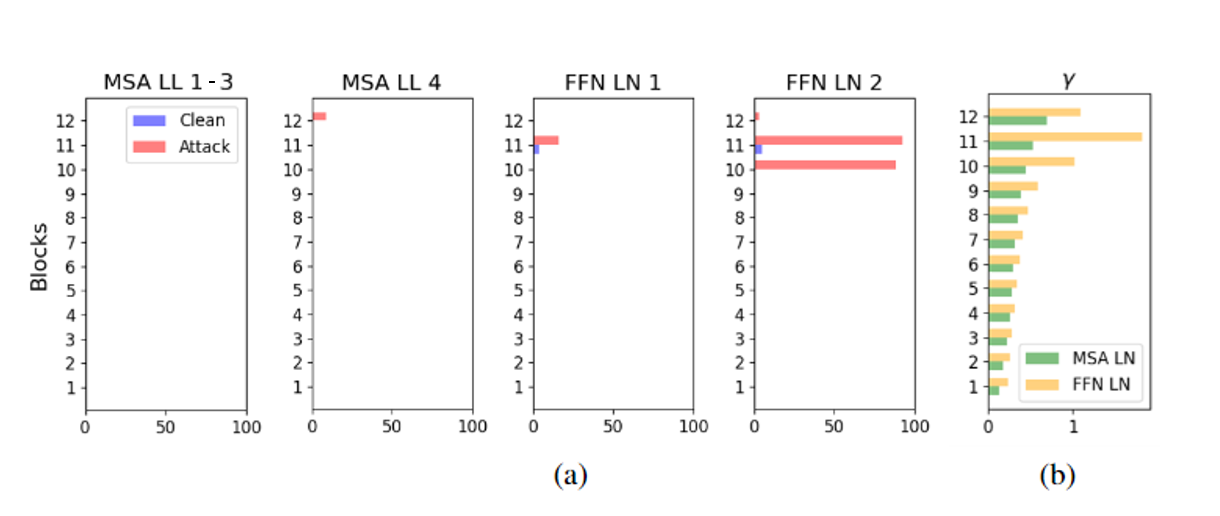}
    \caption{Illustrating the relation between the number of outliers and the normalization layer $\gamma$ parameter. (a) percentage of f16 matrix multiplication across all transformer blocks and linear layers (LL); and (b) the corresponding $\gamma$ values for each normalization layer in each transformer block.
    MSA LL 1-3 are merged for simplicity since there are no outlier values in these layers.
    }
    \label{fig:images}
    \vspace{-0.25cm}
\end{figure}

In Figure~\ref{fig:images}a we show the percentage of outliers in the transformer blocks' six linear layers (explained in Section~\ref{sec:background}).
As can be seen, our attack only affects the number of outliers in blocks 10, 11, and 12.
Particularly, the outliers occur in the 4-th layer of the MSA (\ie, the linear layer that combines the different attention heads) and in both linear layers of the FFN.
We hypothesize that the occurrence of outliers in the MSA 4-th linear layer and in the FFN last linear layer can be attributed to the fact that their inputs are not directly affected by the normalization layers, \ie, additional computations are done between the normalization layers and these layers, allowing our attack to craft synthetic outliers.
In contrast, the inputs to the first three linear layers in the MSA and the first linear layer in the FFN are first processed by the normalization layers.
Interestingly, our attack is not able to affect these MSA layers while successfully generating outliers in the FFN layer.
This phenomena can be explained by examining the $\gamma$ learnable parameter value in each transformer block (theoretically the $\beta$ values should also affect the output's magnitude; however, in practice the values are very low and only marginally affect the magnitude).
As presented in Figure~\ref{fig:images}b, the FFN normalization layer's $\gamma$ value in the first nine blocks is lower than one, scaling down the input values' magnitudes.
However, in the last three blocks the $\gamma$ value is substantially higher, either maintaining or up scaling the input values magnitudes, which contributes to our attack's success.
As opposed to this gradually increasing value pattern, the MSA normalization layer's $\gamma$ value remains low across all blocks, preventing our attack from generating any synthetic outliers that surpass the threshold.
Therefore, we can see a direct correlation between our attack's success and the normalization layer values.

Furthermore, since the normalization layer's output is also affected by the input's mean $\mathrm{E}[x]$ and variance $\mathrm{Var}[x]$, we also tried to attack these values, \ie, decreasing the entire input's mean and variance.
However, it did not improve the performance of our attack.
This arises from the fact that the quantization loss component (Equation~\eqref{eq:quant_loss}) implicitly induces the same effect (we include an analysis of this phenomena in the supplementary material).
Thus, we conclude that the most influential parameter to our attack's performance is the normalization layer's $\gamma$ value.

\vspace{-0.12cm}
\section{\label{sec:countermeasures}Countermeasures}

In response to the challenges posed by QuantAttack, in this section, we propose two practical steps that can be performed to enhance the security of the quantization process:
\textbf{Limiting the use of high-precision multiplications} -- implement rules that limit when and how often the model uses high-precision calculations. 
An implementation of this approach (further discussed in the supplementary material) has shown that the 1681\% increase in the outliers can be reduced to a 251\% increase, without compromising model accuracy on the clean images;
This approach, however, requires a pre-defined threshold which depends on the specific model and system requirements.
\textbf{Increasing the batch size} -- based on our observations (Figure~\ref{fig:batchs}), increasing the batch size reduces QuantAttack's impact due to a higher occurrence of natural outliers.
In this case, the synthetic outliers that our attack produces might blend with the natural outliers (\ie, the number of high-precision multiplications will not increase linearly).
Nonetheless, the quantization's efficiency decreases as the batch size grows.
Above all, both approaches have trade-offs between performance and security, demonstrating that there is no perfect solution that will completely eliminate the threat posed by our attack.

\vspace{-0.15cm}
\section{\label{sec:conclusion}Conclusion}

In this paper, we presented QuantAttack, a novel adversarial attack that both exploits the vulnerabilities and highlights the risk of using vision transformers with dynamic quantization. 
We showed that quantized models, while benefiting from the quantization's efficiency, are susceptible to availability-oriented adversarial techniques that can degrade their performance.
Our comprehensive evaluation demonstrated the impact of our attack on a wide range of vision transformers in various attack configurations.
The implications of these findings are significant, indicating the pressing need to develop more robust quantization methods that can withstand such adversarial challenges.
In future work, we plan to extend our attack to the NLP domain and evaluate its impact on popular large language models (LLMs).

{\small
\bibliographystyle{ieee_fullname}
\bibliography{egbib}
}

\end{document}